%% file: 0_root.tex
\pdfoutput=1
\documentclass[letterpaper, 10 pt, conference]{ieeeconf}  % Comment this line out if you need a4paper

\IEEEoverridecommandlockouts                              % This command is only needed if 
\usepackage{xspace}
\usepackage{siunitx}
\usepackage{times}
\usepackage{epsfig}
\usepackage{graphicx}
\usepackage{amsmath}
\usepackage{amssymb}
\usepackage{gensymb}
\usepackage{siunitx}
\usepackage{dirtree}
\usepackage[font={small}]{caption}
\usepackage{subcaption}
\usepackage{diagbox}
\usepackage{xcolor}
\let\labelindent\relax
\usepackage{enumitem}
\usepackage{hyperref}

\newcommand{\jrdb}{JRDB\xspace}
\newcommand{\jackrabbot}{JackRabbot\xspace}
\newcommand{\methodname}{{JRMOT}\xspace}

\newcommand{\eg}{\emph{e.g.}\xspace}

\newcommand{\etal}{\emph{et al.}\xspace}
\title{\LARGE \bf
\methodname: A Multi-Modal Real-Time 3D Multi-Object Tracker and \\a New Large-Scale Dataset
}

\author{Abhijeet Shenoi, Mihir Patel, JunYoung Gwak, Patrick Goebel, \\\ Amir Sadeghian, Hamid Rezatofighi, Roberto Mart\'in-Mart\'in, Silvio Savarese% <-this % stops a space
\thanks{All the authors are with the Stanford Vision and Learning Laboratory, Stanford University, USA.}%
\thanks{E-mail: {\tt\small[ashenoi,mihirp,jgwak,pgoebel,amirabs,
hamidrt,robertom,ssilvio]@cs.stanford.edu}}%
}

\begin{document}

\maketitle
\thispagestyle{empty}
\pagestyle{empty}

%%%%%%%%%%%%%%%%%%%%%%%%%%%%%%%%%%%%%%%%%%%%%%%%%%%%%%%%%%%%%%%%%%%%%%%%%%%%%%%%
\begin{abstract}
Robots navigating autonomously need to perceive and track the motion of objects and other agents in its surroundings.
This information enables planning and executing robust and safe trajectories.
To facilitate these processes, the motion should be perceived in 3D Cartesian space. However, most recent multi-object tracking (MOT) research has focused on tracking people and moving objects in 2D RGB video sequences.
% This is because searching in 3D is costly, and there are few datasets with 3D sensor modalities annotated with 3D labels of moving agents. 
In this work we present \methodname, a novel 3D MOT system that integrates information from RGB images and 3D point clouds to achieve real-time, state-of-the-art tracking performance.
Our system is built with recent neural networks for re-identification, 2D and 3D detection and track description, combined into a joint probabilistic data-association framework within a multi-modal recursive Kalman architecture. 
As part of our work, we release the \emph{\jrdb} dataset, a novel large scale 2D+3D dataset and benchmark, annotated with over 2 million boxes and 3500 time consistent 2D+3D trajectories across 54 indoor and outdoor scenes. 
\jrdb contains over 60 minutes of data including 360\degree cylindrical RGB video and 3D pointclouds in social settings that we use to develop, train and evaluate \methodname. 
The presented 3D MOT system demonstrates state-of-the-art performance against competing methods on the popular 2D tracking KITTI benchmark and serves as first 3D tracking solution for our benchmark. Real-robot tests on our social robot \jackrabbot indicate that the system is capable of tracking multiple pedestrians fast and reliably. We provide the ROS code of our tracker at \href{https://sites.google.com/view/jrmot}{https://sites.google.com/view/jrmot}
\end{abstract}

%%%%%%%%%%%%%%%%%%%%%%%%%%%%%%%%%%%%%%%%%%%%%%%%%%%%%%%%%%%%%%%%%%%%%%%%%%%%%%%%
\input{1_intro}
\input{2_RelatedWork}
\input{3_Methodology}
\input{4_Data}
\input{5_Experiments}
\input{6_Conclusion}
\bibliographystyle{IEEEtran}
	\bibliography{IEEEabrv,0_root}

\end{document}

%% file: 1_intro.tex
\section{Introduction}
\label{s:intro}

An autonomous agent such as a mobile robot needs to move between two locations in a safe and robust manner. 
To navigate safely, the robot needs to perceive the motion of the multiple dynamic objects and other agents, \eg people and cars, in its vicinity.
This perceived motion allows the agent to predict the possible future trajectories of the other agents and to plan and execute motion strategies that take them into account.

%Why is the problem challenging? Two reasons: 1) detectinga and tracking in 3D is harder than in 2D, 2) there are no datasets in 3D for training DNNs
To facilitate navigation, the motion of the other agents needs to be perceived and represented in the same space the navigation takes place, the 3D Cartesian space.
However, most efforts from the robotics and computer vision communities have been dedicated to the development of multi-object tracking (MOT) systems that perceive 2D motion from RGB video streams. The reason for this is two fold. First, detecting and tracking objects in 3D is computationally more expensive than in 2D due to the \textit{curse of dimensionality} in this search problem. And second, there is a lack of adequate large-scale curated datasets of 3D data with annotations of moving agents from the perspective of navigating robots in human environments, impeding the application of successful deep learning techniques to 3D tracking. 

\begin{figure}[t]
\centering
\includegraphics[width=0.47\textwidth]{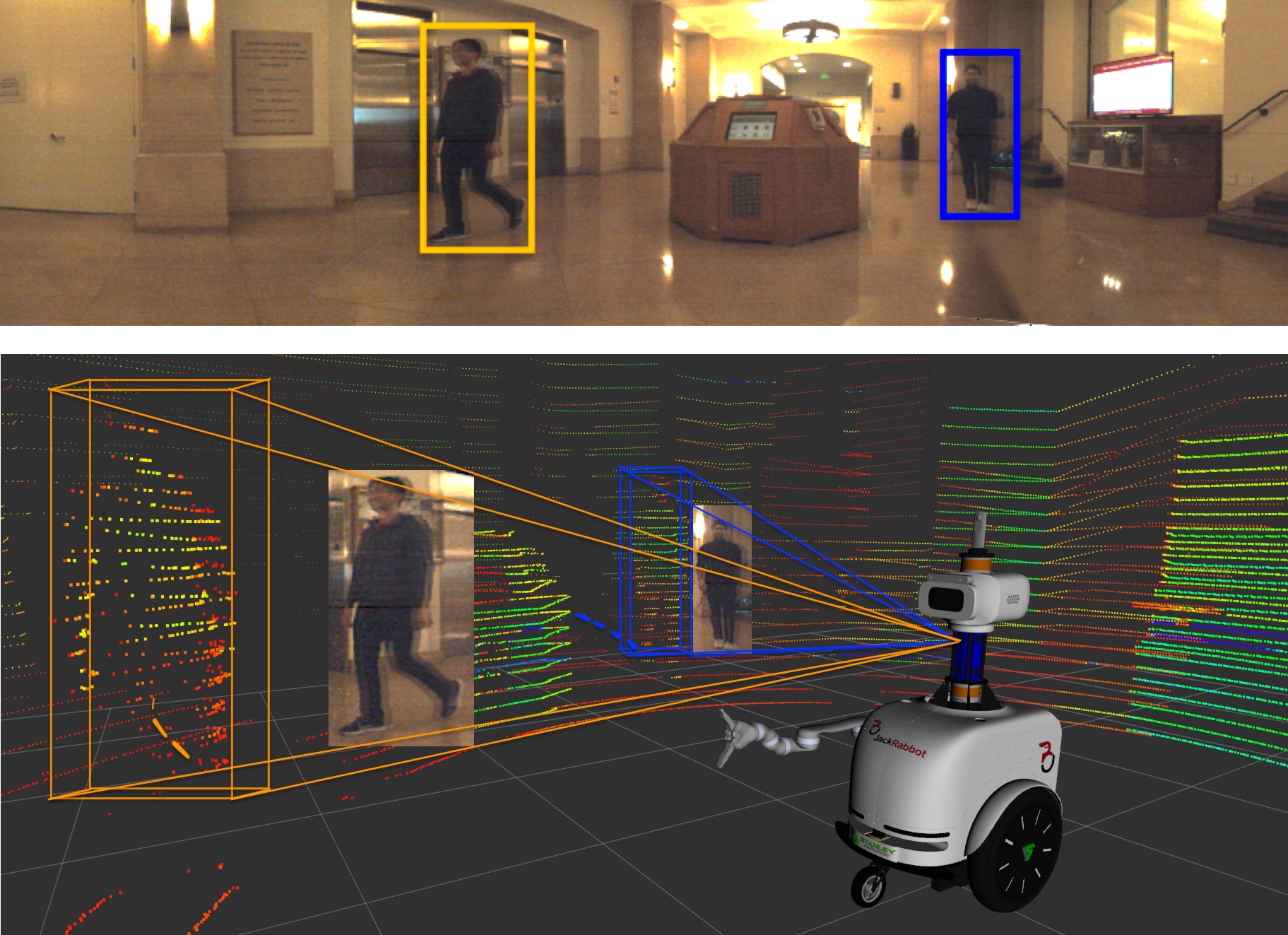}
\caption{Robots navigating in human environments need to detect and track humans and other moving targets using their sensor information; \methodname integrates information from 2D RGB images (top), where appearance is more easily discernible, and 3D point clouds (bottom), where objects are well separated, in a tightly coupled manner to provide real-time 3D multi-object tracking information; \methodname is developed on the \jrdb dataset, a novel annotated dataset captured  with our social mobile manipulator \jackrabbot (bottom)}
\label{fig:pull}
\end{figure}

In this paper we present \methodname, a novel real-time multi-object detection and tracking framework in 3D Cartesian space. \methodname detects and tracks multiple targets around the agent by constraining the 3D search with 2D cues, effectively combining information from RGB cameras and LiDAR sensors. RGB images and 3D pointclouds carry complementary information.
On the one hand, RGB images are \textbf{dense}, which allows us to discern appearances of objects to effectively detect, identify and classify them even at large distances. It is also structured in the form of a \textbf{pixel grid}, well suited to be processed with effective tools such as CNNs. 
On the other hand, 3D point-cloud data is \textbf{sparse} but the depth information allows us to separate objects that might overlap in the 2D image space. However, the unordered structure of the pointclouds do not allow for the use of efficient algorithmic architectures such as CNNs.
\methodname leverages the information of each modality (appearance in RGB, geometry in point clouds) to address the shortcomings of the other by i) sequentially processing them to guide the 3D search in regions indicated by the RGB image, ii) fusing their information into a multi-modal descriptor to facilitate tracking and data-association, and iii) updating the tracking state with a novel multi-modal measurement model.

At its core, \methodname applies state-of-the-art deep neural network architectures to detect objects of interest in RGB images and 3D point clouds, and to characterize tracks with novel multi-modal descriptors, improving the performance of well-established data-association and filtering techniques. Training such networks requires a large amount of 2D RGB images and 3D pointclouds annotated with ground truth labels of the location of objects of interest. The annotated data should be acquired from the perspective of the agent that will execute \methodname, i.e. a mobile robot. \methodname is trained with annotated data from a novel dataset of multi-modal data, the \jrdb dataset. 
The dataset is captured from the perspective of our social autonomous agent, \jackrabbot, leading to a first-of-its-kind dataset that includes \textbf{indoor and outdoor scenes}, with over \textbf{4.2 million annotated bounding boxes} in 2D RGB images and 3D pointclouds.
This dataset enabled us to leverage the complementarity of 2D RGB and 3D pointcloud data with \methodname. 

To summarize, our contributions are:
\setlist{nolistsep}
\begin{enumerate}[leftmargin=*]
\itemsep0em 
    \item We present \methodname, a novel real-time online 3D MOT system that fuses 2D and 3D information based on latest deep-learning architectures. 
    \item We release the \jrdb dataset and benchmark, a first of its kind 2D+3D dataset for the development and evaluation of 2D-3D MOT frameworks and 2D-3D people detection. \methodname is developed and evaluated on \jrdb and serves as first competitive baseline.
    \item We show that \methodname achieves \textbf{state-of-the-art performance} in the competitive KITTI 2D tracking benchmark. Our tests also indicate that our method can detect and track effectively in real time, running onboard a mobile robot, with only few ID switches and a single missed track in over \SI{100}{\second} of experiments. We provide \methodname as ROS code for other researchers to test and build upon.
\end{enumerate}

%% file: 2_RelatedWork.tex
\section{Related Work} 
\label{s:RW}

In recent years, there have been impressive advances in MOT, mostly focused on 2D tracking (in images) with some exceptions of new 3D MOT systems (from images and/or 3D data).
%is comparatively smaller, and most systems share the same components with 2D MOT systems with some minor modifications.
% In this work, we present a 3D MOT system that fuses information from 2D RGB videos and 3D point clouds for tracking multiple objects in 2D/3D space. Moreover, we introduce a novel large-scale dataset to the tracking community for further development and testing of 3D multi-object trackers. This dataset is targeted to a unique visual domain tailored to the perceptual tasks related to navigation in human environments, both indoors and outdoors, and not limited only to self-driving cars but also other types of agents like social mobile robots. 
We now review previous work in the areas of 2D MOT from 2D RGB videos, 3D MOT from 2D RGB and/or 3D sensors, real-time 3D MOT systems and existing datasets for MOT with 3D data.

% Most work in tracking falls into two major paradigms - tracking by detection ~\cite{feng2019multi, henschel2019multiple, DBLP:journals/corr/WojkeBP17, xiangiccv2015, weng2019baseline, yoon2019online}, and joint tracking and detection ~\cite{Luo2018FastAF, DBLP:journals/corr/abs-1710-03958, DBLP:journals/corr/abs-1812-05050, sun2019deep, hu2019joint}. Regardless of whether the task is to track in 2D or 3D, tracking by detection is by far more common and dominates the leader boards on most tracking challenges. However, due to the lack of benchmarks for 2D tracking with available 3D information, and the lack of 3D tracking benchmarks, the focus has primarily been on 2D tracking with access to only 2D sensory inputs.

\textbf{2D MOT with 2D Data:}
Tracking in 2D is the task of perceiving continuously the motion of objects in video sequences. 
There exists a large body of literature for 2D MOT. Works such as~\cite{bergmann2019tracking, feng2019multi, sun2019deep, yoon2019online} leverage the success of deep learning architectures for re-identification, and utilise appearance cues for track-detection association. Other works such as~\cite{wang2019exploit,rezatofighiiccv2015}, use motion and continuity cues to do the same. 
% To alleviate these problems, some approaches attempt to infer 3D properties from only RGB images~\cite{scheidegger2018mono,bertoni2019monoloco}. Inferring 3D properties from 2D images is inherently an ill posed problem that requires additional shape assumptions. 
 \methodname builds on top of this body of literature - we use 2D RGB to obtain appearance descriptors using deep learning based feature extractors, and use 3D pointcloud data to circumvent two problems faced by 2D MOT works. First, the problem of occlusion in 2D is largely reduced due to the large separation of objects in 3D space. Second, motion in 3D is often much more simple than the corresponding projected motion in the 2D RGB image and hence can be used as a much better cue for association. Further, the resulting tracks are in 3D space, which is required by most applications involving autonomous agents.

% Therefore, due to the lack of datasets containing indoor scenes, these approaches are unlikely to work well for autonomous agents that navigate in indoor environments.\roberto{not sure if I buy this last arguments since there is a lot of selfsupervised depth estimation}

\begin{figure*}[th!]
    \centering
    \includegraphics[width=0.85\textwidth]{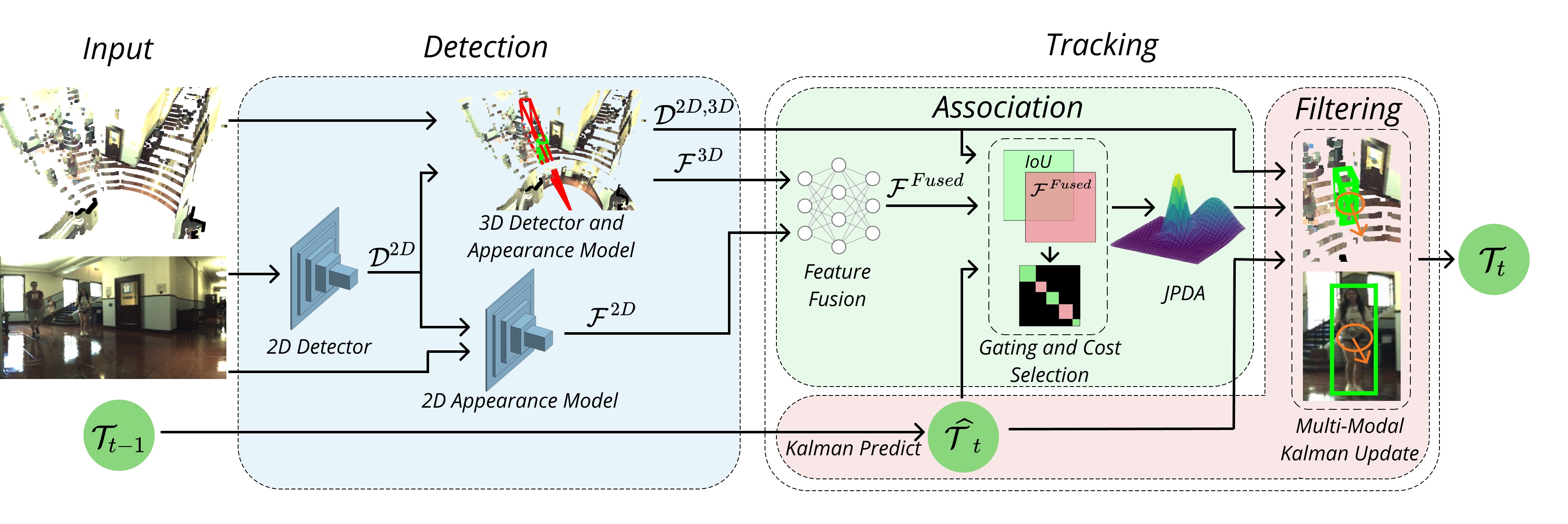}
    \caption{\methodname: Our proposed 3D MOT system is composed of a Detection block, that includes the 2D detector (Sec~\ref{ss:2Ddetection}), 2D appearance model (Sec~\ref{ss:appmod}), 3D detection and feature extractor (Sec~\ref{ss:3Ddetections}), and a Tracking block containing data association (Sec ~\ref{ss:ff} and Sec~\ref{ss:da}) and filtering (Sec~\ref{ss:filtering}) and track management (Sec~\ref{ss:track_manage}) components; $\mathcal{T}, \mathcal{D}, \mathcal{F}$ refer to tracks, detections, features respectively with the superscript indicating the space; The system integrates information from 2D RGB images and 3D pointclouds into a single 3D multi-object recursive estimation tracker with real-time performance}
    \label{fig:framework}
\end{figure*}

\textbf{3D MOT with 2D and/or 3D Data:}
With the advent of self driving cars, access to large scale datasets containing LiDAR data~\cite{caesar2020nuscenes, Geiger2012CVPR, huang2018apolloscape, yu2018bdd100k} has reinvigorated interest in the use of 3D sensors. Some methods use exclusively 3D detections and pointcloud data to perform 3D tracking~\cite{Luo2018FastAF,Weng2019_3dmot}, not using any information from RGB images of the scene. \methodname uses pointcloud data combined with RGB information to improve on tracking of far away objects; this is explored in Fig~\ref{fig:dist_performance}, which shows that as the distance of an object from the robot increases, out method largely mitigates the drop in performance seen in the baseline.

Baser \etal~\cite{baser2019fantrack} aggregate both 2D RGB appearance descriptor and the bounding box coordinates to learn a similarity function to perform 3D tracking. It independently detects in the 2D and 3D domains, and also only utilises 3D measurements to perform filtering, distinct from our work which tightly couples both 2D and 3D measurements. Luiten \etal~\cite{luiten2020track} utilise RGB and depth information to reconstruct a 4D spatio-temporal scene, but unlike our method do so in an offline setting.
% \cite{luber2011people} utilises an RGBD sensor, and uses both 2D and 3D feature descriptors to associate detections to tracks. Here, the measurement used to update the MHT tracker is a purely 3D measurement, utilising 2D information only to perform the association.
The effectiveness of these techniques has \textbf{not been quantitatively tested} because of the lack of a large scale 3D tracking benchmark. They are often evaluated via the proxy of 2D tracking, or on custom generated small-scale datasets.

\textbf{Real-Time 3D MOT Systems}
Whereas there is a plethora of real-time 2D MOT systems such as \cite{wojke2017simple, wang2019towards, sun2019deep} , among many others, the community has developed just a handful of real-time 3D MOT systems. Koide~\etal~\cite{koide2019portable, spinello2011tracking} both propose real time 3D MOT systems based exclusively on 3D LiDAR data without incorporating 2D data. This loose (or complete lack of) coupling of 2D and 3D information is sensitive to cases where 3D detection intermittently fails, whereas the 2D detection is robust, a key advantage of our method as shown in Sec~\ref{ss:res}. Linder~\etal~\cite{linder2016people} and Dondrup \etal~\cite{dondrup2015real} both utilise 2D and 3D data, but do not leverage recent advances in deep learning based detectors and feature descriptors.

\textbf{3D Datasets:} 
3D sensory systems are becoming increasingly commonplace in sensor suites of autonomous agents. Datasets with this multi-modal data such as KITTI~\cite{Geiger2012CVPR}, Apolloscape~\cite{huang2018apolloscape}, NuScenes~\cite{caesar2020nuscenes} and Oxford's Robotic Car~\cite{RobotCarDatasetIJRR} have widely driven research in the 3D community. Nonetheless, their targeted domain of application is autonomous driving; the data they provide is captured from sensor suites on top of cars and only depicts streets, roads and highways. 
Frossard~\etal~\cite{frossard2018end} specifically mention a lack of available 3D tracking benchmarks.
% Additionally, at the time of submission, they have no active 3D tracking challenges.  is a recent dataset and benchmark that has a 3D tracking challenge, but is targeted towards autonomous driving scenarios, in exclusively outdoor settings.

In this paper, we target a unique visual domain tailored to the perceptual tasks related to navigation in human environments, both indoors and outdoors, in crowded scenes. We hope that this new domain provides the community an opportunity to develop visual perception frameworks, limited not only to self-driving cars but also various other types of autonomous navigation agents. Furthermore, we hope this dataset and benchmark will support and drive research in a variety of domains related to social robotics, including but not limited to human detection and tracking.
% , trajectory prediction, and social navigation and planning.

%\fixmer{In summary, 3D tracking of objects, especially pedestrians is an area of research which lacks a reliable benchmark, and there is no currently existing approach which fuses both 2D and 3D information to perform 3D tracking, in an online real-time setting.}

%% file: 3_Methodology.tex
\section{JRMOT: 3D Multi-Object Tracking from 2D and 3D Data}
\label{s:jrmot}

Our proposed 3D MOT system fusing 2D and 3D data is depicted in Fig.~\ref{fig:framework}. \methodname performs tracking by detection. The detector block contains a 2D detector, a 2D appearance feature extractor and a 3D detector (which also generates a 3D feature descriptor). The detector block takes as input a 2D RGB image and the corresponding pointcloud and produces 2D and 3D detections of all objects of interest, along with their 2D and 3D feature descriptors. This is then passed to the tracking block, which performs data association, as well as multi-modal Bayesian filtering. The output of our system is the location in 3D space of all tracked objects, each uniquely identified over time by a track ID. We assume the extrinsic calibration between the RGB camera and the depth sensor to be known. We now explain each component in detail.

\subsection{2D Detection}
\label{ss:2Ddetection}

First, our system needs to detect all moving instances of objects of interest in the environment. Although we are interested on 3D locations, 2D detectors are faster, more robust and accurate than 3D detectors~\cite{8621614}. Therefore, we exploit state-of-the-art image segmentation (Mask R-CNN~\cite{he2017mask}) or object detector (YOLO~\cite{redmon2016you} modified for real time) architectures as our detector. 
The input to this module is a 2D RGB image at time $t$ and the output is a set of $N$ detections in 2D, $\mathcal{D}^{2D}_t = \{(u,v,w,h)_{0},\ldots,(u,v,w,h)_{N-1}\}_t$, where $(u,v)_i$ is the upper-left corner of the detected bounding box around the instance $i$ and $(w,h)_i$ are the width and height of that box. 
The available pretrained models have been trained with different types of images than the ones our robot encounters during navigation. We make use of our \jrdb dataset to finetune the networks and adapt them to the special data distribution of the social navigation setup\footnote{Our detections are publicly released as part of the \jrdb dataset and benchmark for others to use in their MOT systems.}. 

\subsection{2D Appearance}
\label{ss:appmod}

The detections from the previous step need to be associated to existing tracks in \methodname. To this end, we featurize the 2D appearance (appearance in the RGB image) of both detections and tracks in order to compare features and associate them later (Section~\ref{ss:da}).
% Occluded objects are a challenge for any tracking system. 2D appearance can be a useful cue to re-identify the object as explored in ~\cite{DBLP:journals/corr/WojkeBP17} among many other 2D tracking works. 
% While tracking in 3D, the sparse nature of the pointcloud can make it difficult to discern heading direction and lack of color information can make objects of the same shape look very similar.
We compute Aligned-ReID~\cite{zhang2017alignedreid} features when the objects of interest are people, and features from Wu et al.~\cite{wu2018vehicle} when they are vehicles. The choice of these features is based on their high discriminative capabilities and fast computation time.
Both features are trained on \jrdb.
The input to this module is the 2D RGB image at time $t$ and the $N$ detections from the previous step, and the output are their 2D appearance features, $\mathcal{F}^{2D}_t = \{f^{2D}_{0},\ldots,f^{2D}_{N-1}\}_t$.
% For metrics on \jrdb dataset, Aligned ReID was retrained on the train set.

\subsection{3D Detection and Appearance}
\label{ss:3Ddetections}

As mentioned before (Sec.~\ref{s:RW}), it is possible to obtain a noisy estimate of the 3D location of a detected object from its 2D detected box. However, in this work we propose to integrate 2D RGB and 3D data provided by a depth sensor, which is a common part of most autonomous navigating systems. 
We utilise F-PointNet~\cite{DBLP:journals/corr/abs-1711-08488}, a state-of-the-art algorithm to obtain 3D detections in the form of an oriented cuboid around the object instance for every 2D bounding box. F-PointNet estimates a 3D bounding box for that object within the frustum starting at the RGB camera center and passing through the 2D bounding box as illustrated in Fig~\ref{fig:pull}. We choose F-PointNet because it explicitly gives us an association between every 2D and 3D bounding box, it leverages the robustness of 2D detectors, it has a relatively fast inference time, and it has been shown to be one of the top performing 3D detectors on the KITTI benchmark.
% F-PointNet segments out the 3D points within the frustrum that belongs to the target object, and estimates a 3D bounding box for that object.
The input to the 3D detection module is the set of detected 2D bounding boxes around instances of interest at time $t$, $D^{2D}_t$, and the 3D pointcloud at time closest to $t$. 
The output is a set of $M$ detections in 3D for the class of interest at time $t$, $\mathcal{D}^{3D}_t = \{(x,y,z,w,h,l,\theta)_{0},\ldots,(x,y,z,w,h,l,\theta)_{M-1}\}_t$, where $(x,y,z)_j$ is the center of the bottom face of the detected 3D bounding box around the instance $j$, $(w,h,l)_j$ are the width, height and length of that box and $\theta_j$ is the rotation of the box around the normal to the floor plane.

Additionally, we exploit the F-PointNet architecture to generate feature descriptions of the shape of the detected objects, $\mathcal{F}^{3D}_t = \{f^{3D}_{0},\ldots,f^{3D}_{M-1}\}_t$.
The feature from the penultimate layer of F-PointNet is used to regress the 3D bounding box, and thus, it contains information about the 3D shape of the object. We use this feature as a 3D appearance (shape) descriptor.%, fuse it with the 2D appearance descriptor per detection (see next subsection) and use the multi-modal result to improve data association between detections and previous tracked instances.
Note that not every 3D detection has an associated 2D detection. It is possible that F-PointNet does not find a reasonable bounding box within every frustum. Our system accounts for this case, as explained in Sec.~\ref{ss:track_manage}.
% However, in some cases some of the 2D bounding boxes do not lead to a corresponding 3D bounding box due to lack of enough 3D points in the frustrum and so it is possible that the number of 2D bounding boxes and 3D bounding box may be different ($N\neq M$).
% We will overcome this limitation with a multi-modal Kalman filter leveraging both 2D and 3D measurements.

% Since there might be a lack of a 3D detection, due to too few points being within the frustum, not all 2D detections may have an associated 3D detection. The detection contains the center of the bounding box, the dimensions along the coordinate axes, as well as the rotation angle of the box along the vertical axis. The motivation for choosing F-PointNet is two fold. Firstly, it takes advantage of robust 2D detectors, which can detect objects at very large physical distances (even when there are very few LiDAR points corresponding to the object), and secondly, it maintains explicit correspondence between 2D bounding boxes and 3D bounding boxes.

\subsection{Feature Fusion}
\label{ss:ff}

Due to the coupling of 2D and 3D detections, each object of interest now has a 2D feature descriptor, and a 3D feature descriptor. Depending on the conditions (distance, visibility, occlusion) both 2D and 3D appearance can contain valuable information to associate detections and previous tracks. Therefore, we fuse the 2D and 3D features with a 3-layered fully connected network that receives as input $\mathcal{F}^{cat}_t$, given by $\{\left[f^{2D}_0, f^{3D}_{0}\right],\ldots,\left[f^{2D}_{M-1}, f^{3D}_{M-1}\right]\}_t$ where $\left[\right]$ denotes concatenation.
We train this fusion network via metric learning based on a triplet loss and semi-hard negative mining as in Schroff~\etal~\cite{schroff2015facenet}, resulting in a robust feature for association between new detections and previous tracks.
% As a result, the fused feature is close in the learned space for detections at different time steps of the same instance and further for detections of different instances, allowing a robust association between new detections and previous tracks.

\subsection{Data Association}
\label{ss:da}

% \roberto{here we start talking about cost matrices before the reader knows why cost matrices (to do what? why are they needed?) I think rewriting is needed to explain first that we use JPDA on a cost matrices. then explain where they come from. then, explain that jpda on full matrix is costly and there are simplifications (that paragraph of the clusters comes out of the blue and nobody knows what for at that point}

Given a set of detections at time $t$, we need to associate them to tracks at $t-1$ to update the tracks' locations and appearances. To do so we utilize JPDA~\cite{fortmann1983sonar} as it has been shown by~\cite{rezatofighiiccv2015} to be robust to clutter and reduce the occurrence of ID switches. JPDA requires a cost matrix, $C\in \mathbb{R}^{K\times N}$ in which element $c_{ij}$ represents the cost associated with matching track $i$  to detection $j$.
We utilize both appearance and 3D spatial location to associate objects. We first compute appearance similarity by calculating the pairwise $\ell_2$ distance between the $N$ features of detections and the $K$ features of tracks and build an appearance cost matrix, $C_{app}\in \mathbb{R}^{K\times N}$. Then, we compute the location similarity by calculating the pairwise 3D bounding box intersection over union (IoU) assuming that both 3D bounding boxes have the same orientation (same $\theta$), an approximation that generates fairly good results in much shorter computation time. The result is an IoU cost matrix, $C_{IoU}\in \mathbb{R}^{K\times N}$. To simplify the association, we perform \emph{gating} with the Mahalanobis distance (M-distance) with a fixed threshold ($0.95$ quantile from the $\chi^2$ distribution) 

As the size of the cost matrices scales with the square of the number of objects in the scene, association with the entire cost matrix can lead to slow computation. We therefore construct an undirected graph, where every track and detection is a node, and an edge exists between track $i$ and detection $j$ if detection $j$ is within the gate of track $i$. Every connected component in this graph is a cluster. We perform further processing on a per cluster basis leading to a much lower computation time.

Since JPDA requires a single cost matrix representing the association costs, we perform a cost matrix selection (IOU vs. appearance) based on an \textbf{entropy measure}. We select the cost matrix that has a lower entropy per track. A lower entropy cost matrix implies that the cost matrix is more 'peaked', and hence more discriminative.

Given the selected cost matrix, we perform JPDA. To maintain the speed of our tracker, we employ the m-best solution approximation~\cite{Rezatofighi:2016:CVPR} for large clusters. For smaller clusters, complete enumeration is used to obtain the exact solution of JPDA.

To deal with the case where an objects has a 2D detection, but not a corresponding 3D detection, we utilise a two step process. In the first step, all measurements with both 2D and 3D detections are associated with tracks using the procedure above. We then do a second round of cost matrix selection, gating, and JPDA with the appearance cost matrix now only based on the 2D feature descriptor, and the IoU cost calculated with 2D IoU.
% \roberto{within this paragraph there is an important novelty, the entropy measure, but it is just lost into a sequence of "and this, and that" things and it is not motivated why is that even necessary. so basically, this paragraph talks about 2 independent things: clustering in the cost matrices to accelerate performance, and the entropy-based method to select appearance of IoU as source of information for the association. Separate and make clear what you are talking about}

\begin{figure*}[th!]
\centering
\begin{subfigure}[b]{0.27\textwidth}
\includegraphics[width=0.99\textwidth, height=5cm]{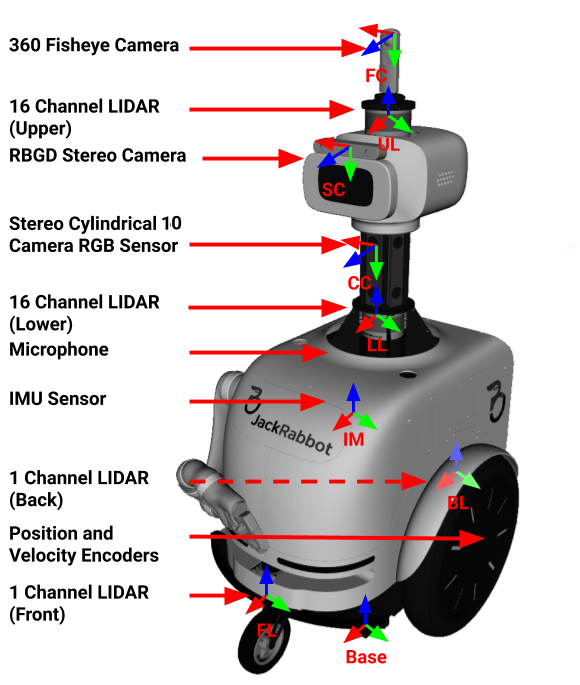}
\caption{\jackrabbot, our data collection platform and its equipped sensors}
\label{fig:JR}
\end{subfigure}
\hfill
\centering
\begin{subfigure}[b]{0.72\textwidth}
\centering
\includegraphics[width=.49\linewidth, height=4.5cm]{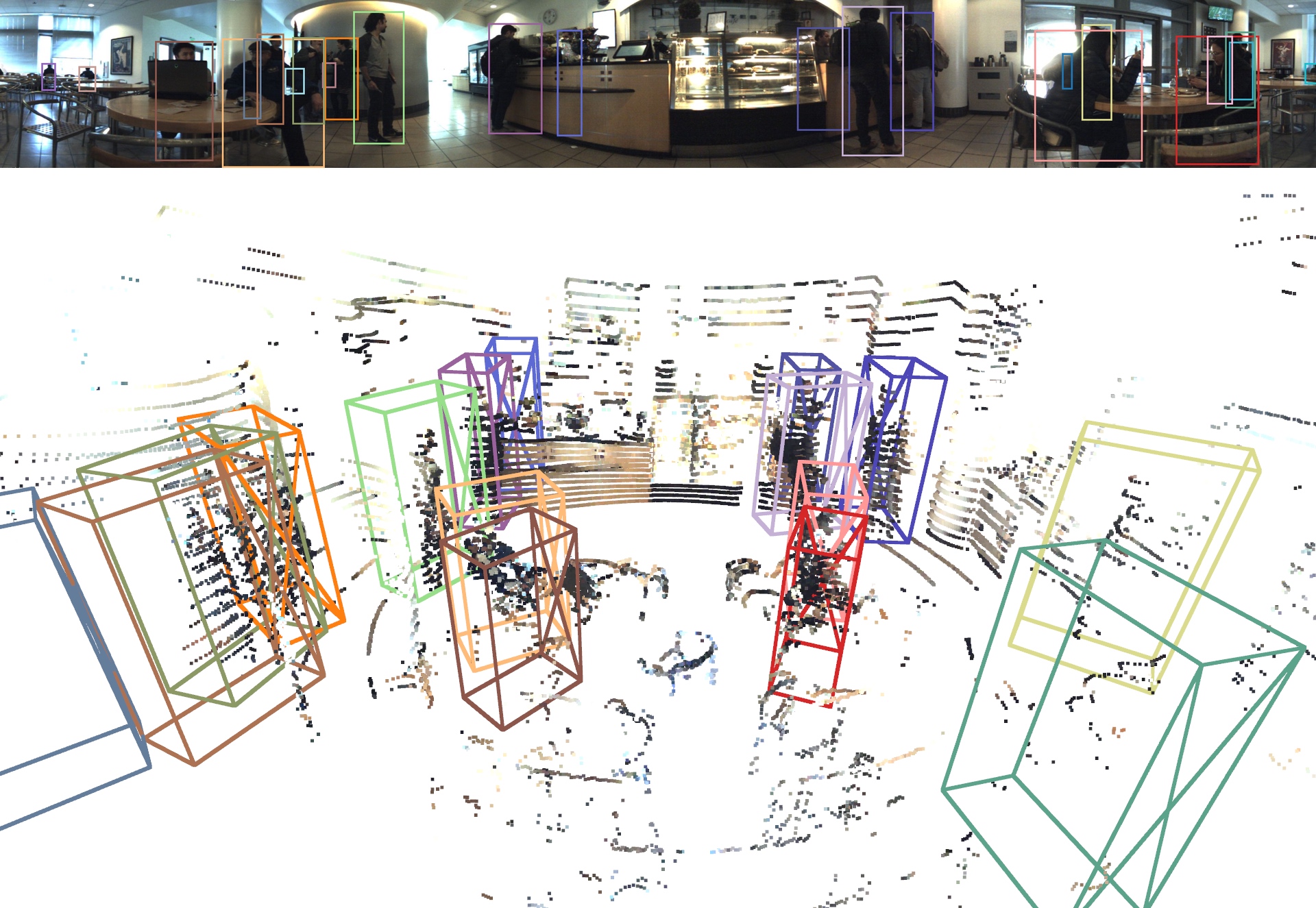}
\includegraphics[width=.49\linewidth, height=4.5cm]{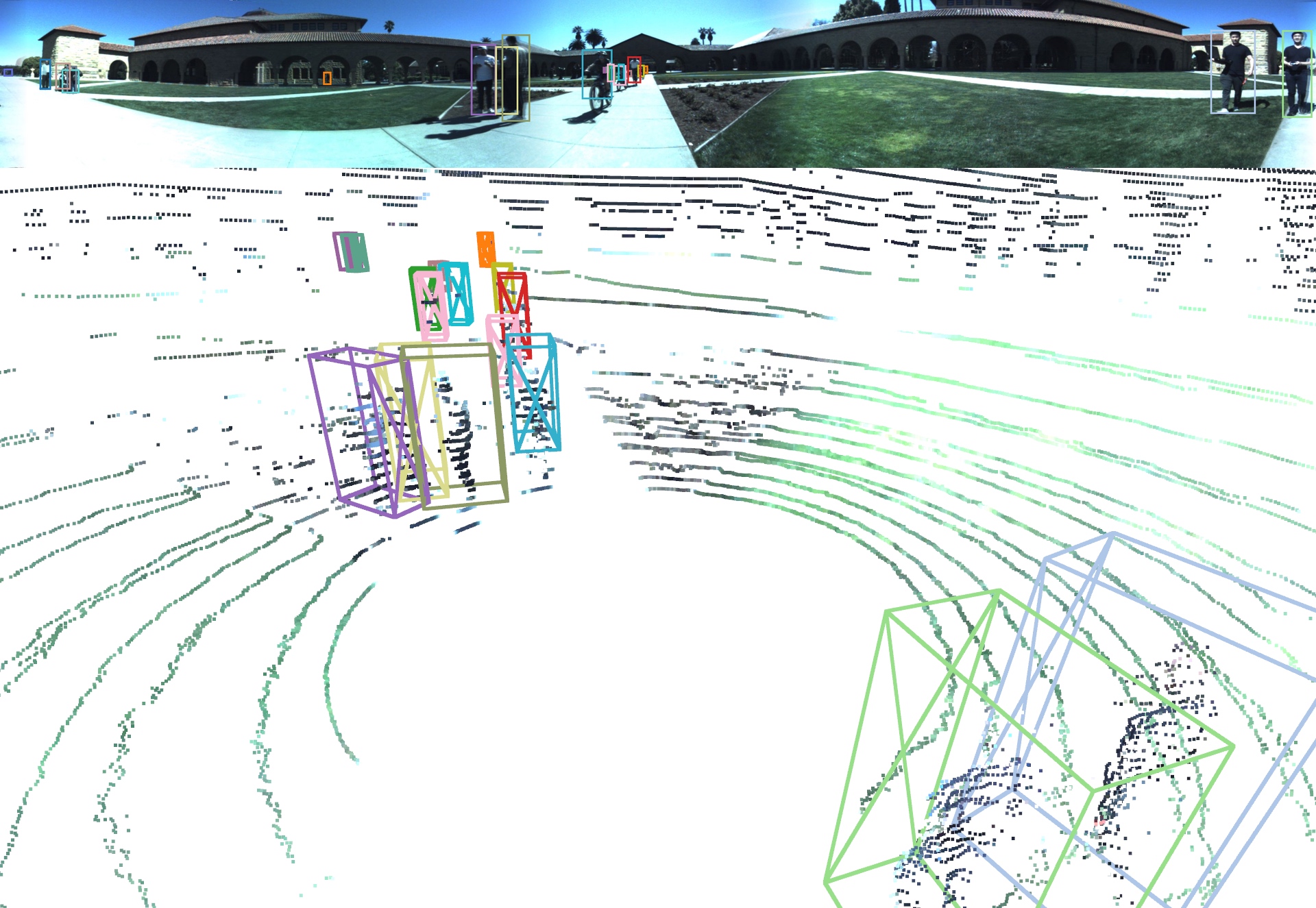}
\caption{Sample visualization of the dataset with stationary (left) and moving (right) robot; Top: 2D stitched 360\degree panorama with human-annotated 2D bounding boxes; Bottom: 3D pointclouds with human-annotated 3D oriented bounding boxes; 2D and 3D annotations have same IDs (indicated by similar box color)}
\label{fig:visualization}
\end{subfigure}
\vspace{-10pt}
\end{figure*}

\subsection{Filtering}
\label{ss:filtering}

2D and 3D detections are often noisy. Therefore, we filter them over time with a Kalman filter~\cite{kalman1960new} to estimate smooth 3D tracks. The Kalman filter is an optimal estimator (assuming Gaussian noise and linearity in motion) and an online, computationally efficient process that allows \methodname to be accurate and real time.

The state to estimate per object includes its 3D location, $x, y, z$, its dimensions approximated as a 3D bounding box, $l, w, h$ and the rotation of the box about the vertical axis, $\theta$. Since objects in most scenes move along the horizontal axes, $X$ and $Z$, and with very small variation on their orientation, we only track the velocities along $X$ and $Y$ axes, $v_x$ and $v_y$ respectively. Hence, the state of each object $O$ is $\bar{\mathbf{x}^{O}} = \{x, y, z, l, h, w, \theta, v_x, v_z\}^{O}$. We apply an independent Kalman filter for each object with a constant velocity motion model for predictions.

To leverage the joint nature of the detections and the multi-modal (2D and 3D) sensor source, we use a \textbf{dual measurement update}. Each track has two measurement sources, the 2D bounding boxes, as well as the 3D bounding boxes that we assume to be independent, although this is not strictly the case. We combine a first PDA~\cite{bar2009probabilistic} Kalman filter update based on 3D measurements, with a second PDA Extended Kalman Filter (EKF) update, with the 2D measurements. The first linear measurement update serves as the primary component and carries most information, with the 2D measurement acting as a fine tuning measurement correction.

For those 2D detections without a corresponding 3D detection, we perform a PDA update of the tracks with the 2D measurement only.

\subsection{Track Management}
\label{ss:track_manage}

% The discussion so far has revolved around the assumption that we have tracks at time $t-1$ and after performing association, we update the track state to account for measurements at time $t$. However, 
\textit{Creating and terminating tracks:} When a new object enters the scene, a new track is initiated only if it is outside the gate of all existing tracks. In that case, we create a temporary track (not part of the \methodname output) and only after $n_{init}$ number of consecutive matches, we promote it to full track. This process reduces noise and avoids false positives. Further, we terminate a track if there has been no matching detection for $n_{term}$ consecutive frames, to account for objects leaving the scene. 
% \roberto{these constants need to be define later in exp evaluation, in case they are not. usually, you can have a 1-2 sentence paragraph defining all params}

\textit{Updating tracks' appearance:} At each step, we update the appearance of the tracks with the latest RGB and pointcloud information to facilitate association in the next step. To do so, we need to associate each last detection to only one track. However, JPDA provides a full probability distribution of association between tracks and detections. Therefore, we perform a linear sum assignment on the JPDA output using the Hungarian algorithm~\cite{Kuhn55thehungarian} with $p_{assn}$ as the minimum probability for a match to be considered. This process provides one-to-one associations that allow us to update the feature descriptor of each track with the assigned best detection. If for a track no detection is assigned, its features are not updated.

%% file: 4_Data.tex
\section{Dataset}

As reviewed in Sec.~\ref{s:RW}, datasets with the type of 3D data and annotations necessary to develop and train 3D MOT systems are scarce and focused on autonomous driving scenarios: there is a need for novel datasets with 3D annotations in social environments from the perspective of navigating robots. We present the \textbf{\jrdb dataset}, a novel dataset focused on human social environments. Our dataset contains {\bf 64 minutes} of sensor data acquired from our mobile robot \jackrabbot comprising {\bf 54 sequences} indoors and outdoors in a university campus environment. In this section, we summarize the data collection and labeling process of the dataset. % For full details, please refer to the supplementary material.

%\url{http://svl.stanford.edu/projects/jackrabbot/}

\subsection{The \jackrabbot Social Robot}

% \begin{figure}[h!]
%     \centering
%     \includegraphics[width=0.45\textwidth]{figs/jr_angle_blur.png}
%     \caption{\jackrabbot, our data collection platform, is equipped with 4 LIDAR sensors, 3 cameras, 2 movement encoders, an IMU sensor, and a microphone.}
%     \label{fig:JR}
% \end{figure}

% \begin{figure*}[t]
% \centering
% \begin{subfigure}[b]{0.48\textwidth}
%     \centering
% 	\includegraphics[width=.9\linewidth]{figs/jrdb_viz_00600.png}
% 	\caption{\textit{bytes-cafe-2019-02-07\_0}: Stationary robot at indoor scene.}
% \end{subfigure}
% \hfill
% \begin{subfigure}[b]{0.48\textwidth}
%     \centering
% 	\includegraphics[width=.9\linewidth]{figs/jrdb_viz_17200.png}
% 	\caption{\textit{memorial-court-2019-03-16\_0}: Moving robot at outdoor scene.}
% \end{subfigure}
%  	\caption{Sample visualization of the dataset. Top: 2D stitched 360\degree panorama with human-annotated 2D bounding boxes. Bottom: 3D Velodyne point clouds with human-annotated 3D rotated bounding boxes.}
%  	\label{fig:visualization}
% \end{figure*}

The \jrdb is a multimodal dataset collected with the sensors on-board of our mobile manipulator \jackrabbot. \jackrabbot is a custom-design robot platform tailored to navigate and interact in human environments. It is equipped with a state-of-the-art sensor suite including stereo RGB 360\degree cylindrical video streams (resulting from composing images from to rows of five aligned cameras each), 3D point clouds from two 16 lines LiDAR sensors, front and back single line LiDAR pointclouds, RGB-D and 360\degree spherical RGB images from the cameras on the head, audio, IMU and GPS sensing. Fig~\ref{fig:JR} depicts \jackrabbot and its on-board sensors. Our goal is to investigate and develop novel solutions for perception and high-level social interactions between humans and robots through \jackrabbot.

\subsection{Data Collection and Annotation}

To generate \jrdb, we collected data in 30 different locations indoors and outdoors, all in a university campus environment, with varying and uncontrolled environmental conditions such as illumination and other natural and dynamical elements. We also ensure the recorded data captures a variation of natural human posture, behaviour and social activities in different crowd densities. Furthermore, to incorporate a diversity in the robot's ego-motion, we use a combination of static and moving sensor (robot) views to capture the data.

A crucial component in the development of social autonomous navigating agents is to perceive and understand the location and motion of humans surrounding the robot. Therefore, in this first round of annotation we focus on detecting and tracking humans. We include the following ground truth labels in \jrdb: a)~over 2.4 million 2D bounding boxes for human/pedestrian class in both the ten separate RGB images and the two composed cylindrical 360\degree images, b)~over 1.8 million 3D oriented bounding boxes for human/pedestrian class in pointclouds from the two 16-lines LiDAR sensors, c)~spatial ID association between corresponding 2D and 3D bounding boxes (all 3D boxes have an associated 2D box but not vice versa), and d)~temporal ID association with time consistent identities for all annotated pedestrians in both 2D and 3D. Fig~\ref{fig:visualization} depicts examples of \jrdb and the annotated ground truth labels on both an RGB 360\degree cylindrical image and a 3D LiDAR pointcloud, colored with the information from the RGB image. With this unique dataset, we hope to facilitate and enable novel research in social navigating autonomous agents. We will augment \jrdb in the future with additional annotations related to social understanding in human environments such as 2D human skeleton posture and individual, group and social activity.

%% file: 5_Experiments.tex
\vspace{-0.025cm}
\section{Experimental Evaluation}
\label{s:exp}

We adopt the standard Clear-MOT metrics~\cite{bernardin2008evaluating} in our evaluation, including accuracy (MOTA), precision (MOTP), and number of ID switches (IDS) along with runtime, as we aim at developing an online real-time MOT system. However, Clear-MOT metrics were developed for 2D tracking, \eg tracks are designated true of false positives based on IoU between estimated and ground truth 2D bounding box in the RGB images. We extend these definitions to 3D based on 3D IoU computed combining the Sutherland-Hodgman algorithm~\cite{sutherland1974reentrant} and Gauss's area formula to determine the volume of the intersection.

% We do not test on the NuScenes dataset because annotations are only at 2Hz, and out method is intended to be a real time online system
%The evaluation in KITTI allows us to compare to a large corpus of existing MOT systems, while our results on \jrdb act as a competitive first set of results for 3D people tracking in this novel and challenging domain. 

Our goal is to develop a real time online MOT system for navigating robots in human environments. Therefore, we evaluate \methodname on our novel \jrdb dataset (3D) and the well-established KITTI dataset~\cite{geiger2013vision}.
The KITTI dataset contains 2D RGB images and 3D pointclouds, but the benchmark only reports 2D tracking results with 2D Clear-MOT metrics. 
% Although our goal is to develop a 3D tracker, we think it is valuable to compare our system to previous approaches on KITTI. 
% Therefore, we measure the performance of our 3D tracker on the 2D tracking KITTI benchmark by evaluating the quality of our 2D tracks. 
Though \methodname is a 3D MOT system, evaluating on KITTI allows us to compare to existing tracking methodologies.
To be able to evaluate \methodname on KITTI, we modify the system presented in Sec.~\ref{fig:framework} by changing the state in our filter architecture to $\{x,y,w,h,v_x,v_y\}$, 
where $x,y,w,h$ parameterize the 2D bounding box and $v_x, v_y$ give the velocity in the 2D image.
% As it is standard in KITTI, we evaluate 2D tracking of cars and pedestrians separately.
% While 2D tracking is not the objective of this framework, strong results would reinforce the belief that 2D and 3D tracking are inherently linked and demonstrate the advantages of our framework.\roberto{this should come to the analysis, because it talks about results}
The \jrdb dataset and benchmark contains both RGB and pointcloud inputs, groundtruth 3D bounding boxes of pedestrians and an evaluation script for 3D tracking, which we use. We compare the results of \methodname to a state-of-the-art baseline, AB3DMOT~\cite{Weng2019_3dmot}, on people tracking. We choose AB3DMOT as baseline due to it being a real time, online tracker, and the availability of the open-source code. At the time of submission, no other open-source online 3D MOT systems were available.

In order to provide comparable results, we aim to use identical detection inputs for all methods. For the KITTI dataset, we only use publicly available detections for the car and pedestrian challenges. For \jrdb, we use the same set of  Mask-RCNN detections for all methods.
% While this puts us at a disadvantage compared to methods that use more recent detectors on KITTI, we believe this is the only way to ensure meaningful results that only reflect tracker performance.
The publicly available detections for KITTI we chose were RRC~\cite{DBLP:journals/corr/RenCLSPYTX17} detections for cars and SubCNN~\cite{7926691} detections for pedestrians.
For our evaluation of AB3DMOT on \jrdb, which requires 3D detections as input, we used the 3D detections from F-PointNet which were generated as a by product from our tracking system.

For experiments on KITTI we use the following parameter settings: $p_{assn}=0.65,n_{init}=2,n_{term}=2$. For experiments on JRDB, we use: $p_{assn}=0.6,n_{init} = 3,n_{term}=5$.
\subsection*{Results}
\label{ss:res} 

% \begin{table}[tbh]
% \footnotesize
%     \centering
% \begin{tabular}{ | c | c | c | c | c |}
% \hline
%  & MOTA $\uparrow$ & MOTP $\uparrow$ & MT $\uparrow$ & Runtime $\downarrow$\\ 
%   \hline\hline
%  MASS~\cite{8782450} & 85.04\% & 85.53\% & \bf 74.31\% & 0.01s\\
%  \hline
%  mmMOT~\cite{luiten2019track} & 84.77\% & 85.21\% & 73.23\% & 0.01s \\ \hline
%  3DT~\cite{hu2019joint} & 84.52\% & 85.64\% & 73.38\% & 0.03s \\ \hline
%  MOTBP~\cite{8461018} & 84.24\% & \bf 85.73\% & 73.23\% & 0.3s \\ \hline
%  IMMDP~\cite{Xiang_2015_ICCV} & 83.04\% & 82.74\% & 60.62\% & 0.19s \\ \hline
%  aUToTrack~\cite{DBLP:journals/corr/abs-1905-08758} & 82.25\% & 80.52\% & 72.62\% &  0.01s \\ \hline
%  JCSTD~\cite{8621602} & 80.57\% & 81.81\% & 56.77\% & 0.01s \\
%  \hline\hline
%  Ours & \bf 85.70\% & 85.48\% & 71.85\% & 0.07s \\
% \hline
% \end{tabular}
%     \caption{Results on Online KITTI Car Tracking}
%     \label{tab:kitti_car_table}
%     \vspace{-5pt}
% \end{table}

\begin{table}[t]
\vspace{0.15cm}
\footnotesize
    \centering
\begin{tabular}{ | c | c | c | c | c |}
\hline
 & MOTA $\uparrow$ & MOTP $\uparrow$ & IDS $\downarrow$ & Runtime $\downarrow$\\ 
   \hline\hline
 MASS~\cite{8782450} & 85.04\% & 85.53\% & 301 & 0.01s\\
 \hline
 mmMOT*~\cite{luiten2020track} & 84.77\% & 85.21\% & 284 & 0.01s \\ \hline
%  3DT~\cite{hu2019joint} & 84.52\% & 85.64\% & 377 & 0.03s \\ \hline
 MOTBP*~\cite{8461018} & 84.24\% & \bf 85.73\% & 468 & 0.3s \\ \hline
 IMMDP~\cite{Xiang_2015_ICCV} & 83.04\% & 82.74\% & 172 & 0.19s \\ \hline
%  aUToTrack~\cite{DBLP:journals/corr/abs-1905-08758} & 82.25\% & 80.52\% & 1025 & 0.01s \\ \hline
 JCSTD~\cite{8621602} & 80.57\% & 81.81\% & \bf 61 & 0.01s \\
 \hline\hline
 Ours* & \bf 85.70\% & 85.48\% & 98 & 0.07s \\
\hline
\end{tabular}
    \caption{Results on online KITTI car tracking benchmark. *~indicates that the method used the same public detections as our method}
    \label{tab:kitti_car_table}
    \vspace{-5pt}
\end{table}

% \begin{table}[tbh]
% \footnotesize
%     \centering
% \begin{tabular}{ | c | c | c | c | c |}
% \hline
%  & MOTA $\uparrow$ & MOTP $\uparrow$ & MT $\uparrow$ & Runtime $\downarrow$\\ 
%   \hline\hline
%  CAT~\cite{nguyen2019ISPRS} & \bf 52.35\% & 71.57\% & \bf 34.36\% &  \textit{Not Reported}\\ \hline
%  Be-Track~\cite{Dimitrievski2019sensors} & 51.29\% & \bf 72.71\% & 20.96\% & 0.02s\\
%  \hline
%  MDP~\cite{Xiang_2015_ICCV} & 47.22\% & 70.36\% & 24.05\% & 0.9s \\ \hline
%  JCSTD~\cite{8621602} & 44.20\% & 72.09\% & 16.49\% & 0.07s \\ \hline
%  SCEA~\cite{Yoon_2016_CVPR} & 43.91\% & 71.86\% & 16.15\% & 0.06s \\ \hline
%  RMOT~\cite{7045866} & 43.77\% & 71.02\% & 19.59\% & 0.02s \\ \hline
%  AB3DMOT~\cite{Weng2019_3dmot} & 36.36\% & 64.86\% & 14.09\% & 0.0047s \\ 
%  \hline\hline
%  Ours & 45.98\% & 72.63\% & 23.02\% & 0.06s \\
% \hline
% \end{tabular}
%     \caption{Results on Online KITTI Pedestrian Tracking}
%     \label{tab:kitti_ped_table}
%     \vspace{-5pt}
% \end{table}

\textbf{KITTI Dataset:} Table~\ref{tab:kitti_car_table} shows our results in the car tracking challenge. We achieve state-of-the-art performance (highest MOTA) among all online published 2D MOT methods. 
%We show competitive results (top 5) when considering MOTP for all online real time published methods. 
Our MOTP is within 0.5\% of the leader and we are second in terms of ID switches and beat all other top submissions by sizable margins.

Table~\ref{tab:kitti_ped_table} shows our results in the pedestrian tracking challenge. Amongst competing real-time methods (computation time less than $0.1s$), our tracker ranks second.

Only one other method uses the same detections as our method. We remain within 1.5\% MOTA, while running in only $\frac{1}{15}^{th}$ the time.
The performance gains in our method are a consequence of fusing and fully leveraging complementary information in 2D RGB and 3D pointcloud information. 
% \textcolor{red}{We hypothesize low detection thresholds combined with better updates give us decreased FP and FN, but result in fragmented trajectories for highly occluded parts other trackers might ignore, yielding higher IDS. This is best evidenced by our 2nd place ranking in minimizing mostly lost sequences}{} \textcolor{orange}{(not displayed here)}\textcolor{red}{.}
One point to note is the higher IDS of \methodname. We found that optimizing MOTA decreases FN's, at the expense of higher IDS. This hyperparameter optimization is only specific to KITTI pedestrian tracking as evidenced by Table~\ref{tab:kitti_car_table}, where our method achieves the $2^{nd}$ lowest IDS.
Even though our method was developed for 3D MOT, \methodname ranks among the state-of-the-art 2D MOT systems in KITTI benchmark, indicating the benefits of our proposed approach, and validating the effectiveness of the system.
% However, using the publicly available detections on KITTI hinders our method since better detections lead to better tracking performance, as suggested by the fact that in the pedestrian benchmark all top methods utilize private detections from more modern detectors.
\begin{table}[th!]
\footnotesize
    \centering
\begin{tabular}{ | c | c | c | c | c |}
\hline
 & MOTA $\uparrow$ & MOTP $\uparrow$ & IDS $\downarrow$ & Runtime $\downarrow$\\ 
   \hline\hline
 CAT~\cite{nguyen2019ISPRS} & \bf 52.35\% & 71.57\% & 206 &  \textit{Not Reported}\\ \hline
 Be-Track~\cite{Dimitrievski2019sensors} & 51.29\% & \bf 72.71\% & 118 & 0.02s\\
 \hline
 MDP*~\cite{Xiang_2015_ICCV} & 47.22\% & 70.36\% & 87 & 0.9s \\ \hline
 JCSTD~\cite{8621602} & 44.20\% & 72.09\% & \bf 53 & 0.07s \\ \hline
%  SCEA~\cite{Yoon_2016_CVPR} & 43.91\% & 71.86\% & 56 & 0.06s \\ \hline
 RMOT~\cite{7045866} & 43.77\% & 71.02\% & 153 & 0.02s \\ \hline
 AB3DMOT~\cite{Weng2019_3dmot} & 36.36\% & 64.86\% & 142 & 0.0047s \\ 
 \hline\hline
 Ours* & 45.98\% & 72.63\% & 395 & 0.06s \\
\hline
\end{tabular}
    \caption{Results on Online KITTI Pedestrian Tracking. *~indicates that the method used the same public detections}
    \label{tab:kitti_ped_table}
    \vspace{-5pt}
\end{table}

\textbf{\jrdb Dataset:} 
\methodname outperforms the baseline, AB3MOT, on the \jrdb benchmark with \textbf{20.2\%} MOTA at \textbf{25 fps} (compared to 19.3\% MOTA of AB3MOT). These MOTA values indicate that the scenes in our dataset are extremely challenging and will guide new research in the field. Based on the $765,907$ false negatives of our method on the test set, we infer that 3D detections are the limiting factor in our tracking system. 

\begin{figure}[h]
\centering
\begin{subfigure}{0.23\textwidth}
\includegraphics[width=0.99\textwidth]{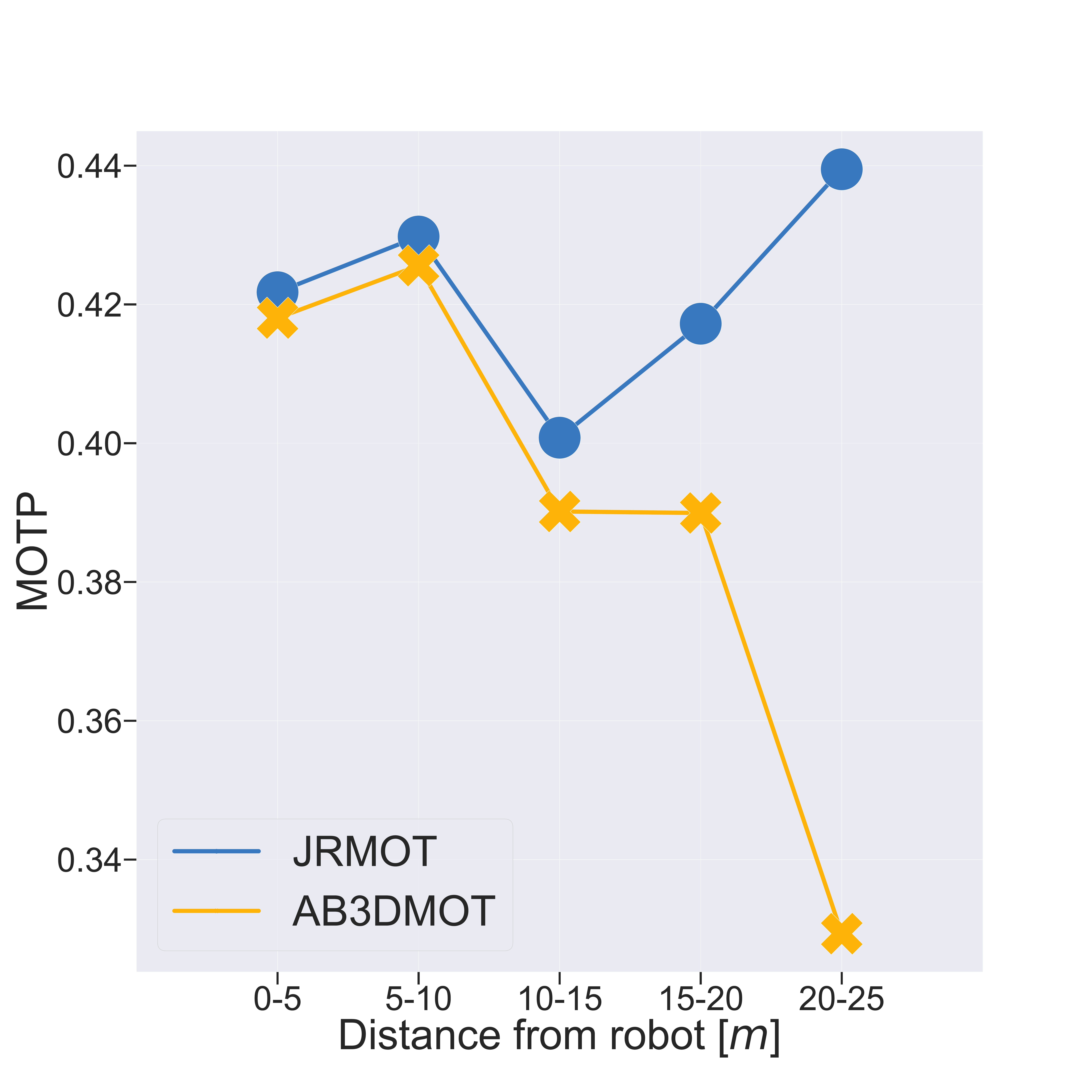}
\caption{}
\end{subfigure}
\centering
\begin{subfigure}{0.23\textwidth}
\centering
\includegraphics[width=.99\textwidth]{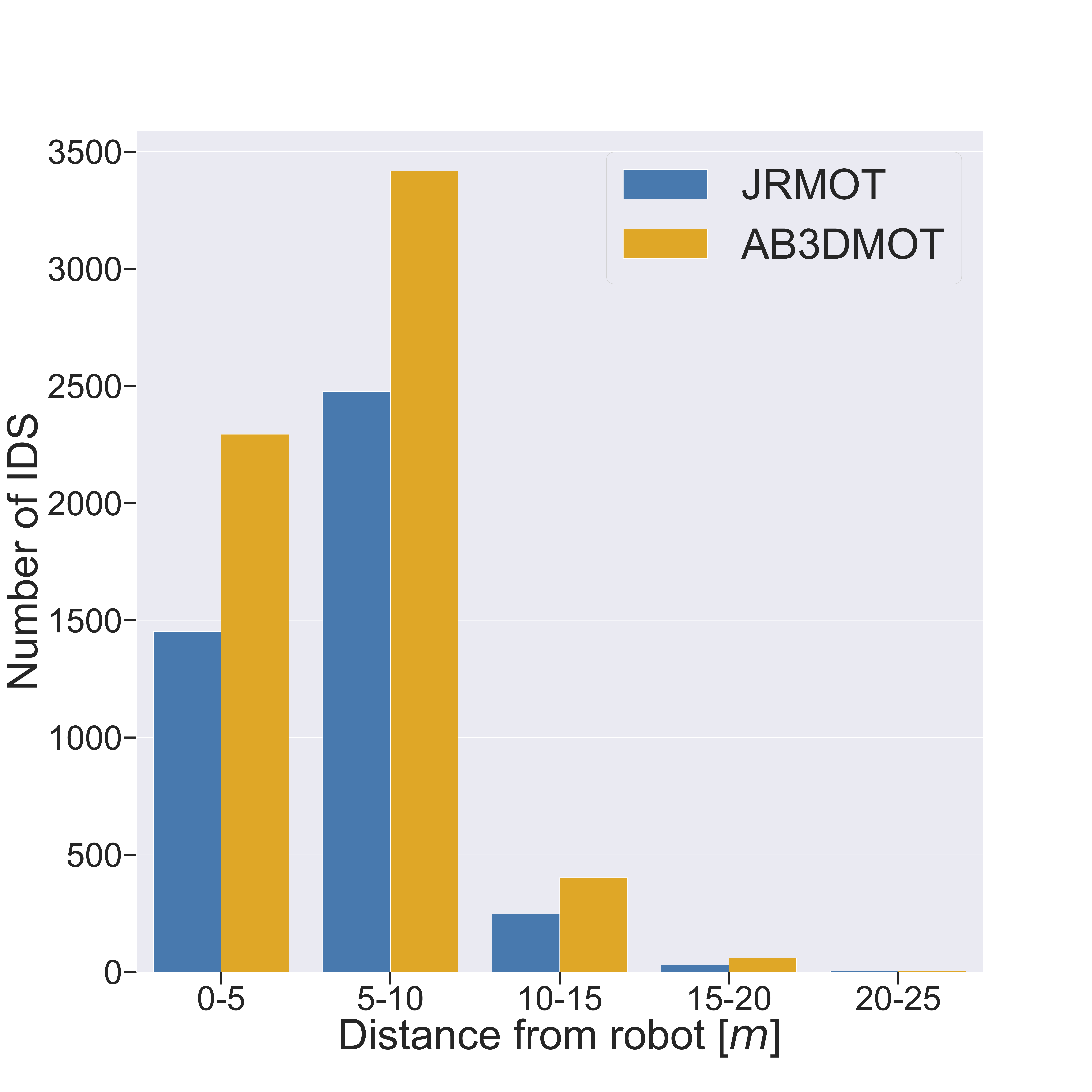}
\caption{}
\end{subfigure}
\caption{Comparison of \methodname and AB3DMOT as a function of distance. a) \methodname obtains higher MOTP due to a more accurate estimation of the orientation of boxes and fine grained position information through all distances b) Our method also has less IDS (lower is better) than AB3DMOT, indicating more robust and stable tracking at all distances}
\label{fig:dist_performance}
\end{figure}

The benefits of the \methodname approach to combine 2D and 3D information are clearer for tracks relatively further away from the sensor, where 3D pointcoud data is sparse, but 2D RGB is a rich source of information. To verify this, we analyze the results as a function of the distance between tracks and robot. Although the MOTA remains fairly similar across all distances, with our method outperforming the baseline, we make the following observations. First, we observe that our hypothesis that 2D data is useful to improve orientation of 3D bounding boxes and make fine adjustments to position is validated in Fig.~\ref{fig:dist_performance}. It can be seen that as the distance from the robot increases, the MOTP of AB3DMOT degrades considerably, whereas our method is consistent across all distance ranges. Further, our method has 30\% fewer ID switches. This shows that our method is able to assign a consistent track ID to individual people, far better than AB3DMOT, across all distances. 
% Further and observed that \methodname outperforms clearly the baseline at the range between \SI{15}{\meter} and \SI{20}{\meter} with nearly 9\% MOTP improvement. This results indicates that in \methodname the primary measurement update for track location is based on the 3D data, while the 2D RGB information serves to estimate the orientation and make small corrections to the location of the 3D bounding box; in the range between \SI{15}{\meter} and \SI{20}{\meter} this strategy provides the best results, while if closer, the update based on 3D pointcloud dominates. 
% This seemingly counter intuitive fact can be due to the following: Even though 2D IoU is not a reliable association metric, 3D IoU is extremely good. However, there are a few cases where 3D IoU on its own cannot effectively accomplish the data association. We hypothesise our improved results are due to our fusion of both 2D and 3D measurements as confirmed by our ablation studies.

Additionally, we analyse the contribution of the individual components in the overall performance of \methodname on a set of ablation studies on the \jrdb dataset. First, we conduct an experiment where we update the tracks only with 2D measurements. As expected, we observe that the 3D information is the most crucial for 3D tracking: without 3D data we obtain -20.1\% MOTA on the train set. We also analyse the contribution of the 2D RGB appearance feature by using only 3D IoU as association metric. In this case, we see a small degradation in performance of 0.1\% MOTA. This indicates that the \textbf{3D IoU is the most informative association metric}, but it is slightly improved in some cases with 2D appearance. Our last ablation is to verify that 2D inputs without corresponding 3D bounding boxes are indeed helpful measurements in our MOT system. We observed that if we do not use these only-2D updates, the MOTA remains constant at 42.9\% but the MOTP drops 0.6\%.
The overall contribution of 2D is therefore 0.1\% MOTA and 0.6\% MOTP. However, this is misleading, due to the large number of objects that are close to the sensor, where 2D information is not expected to help much. \textbf{In the 15-20m range, the increase by using 2D information (both appearance and measurements) is 1.3\% MOTA}. This confirms our intuition that the 2D measurement can be used to make fine updates on the tracked orientation and location, especially further away from the robot. % Visualizations of the resulting 3D tracks with our method and the ablated versions compared to ground truth are included in the supplementary material.
% \fixmea{Write about JR results. What does this indicate? Ideas: Baseline is close bc 3D IoU is very informative and same detection set. We are better by 2D update? appearance? Look at ablation?}

\section{Real Robot Evaluation}
\label{s_rre}

Finally, we evaluate the performance of \methodname when running on-board of a real robot platform. We test on our social robot \jackrabbot, which was used to collect the \jrdb dataset. We chose not to run \methodname at the same time as we collected all data for the \jrdb dataset as it is not possible due to computing limitations (recording images and pointclouds considerably slows down tracking performance). Therefore, we cannot compute MOTA and MOTP on annotated data while running in real-time on the robot; we instead analyze the number (ID switches), as well as the number of lost tracks.

\begin{figure}[h]
\centering
\begin{subfigure}{0.15\textwidth}
\includegraphics[width=0.99\textwidth]{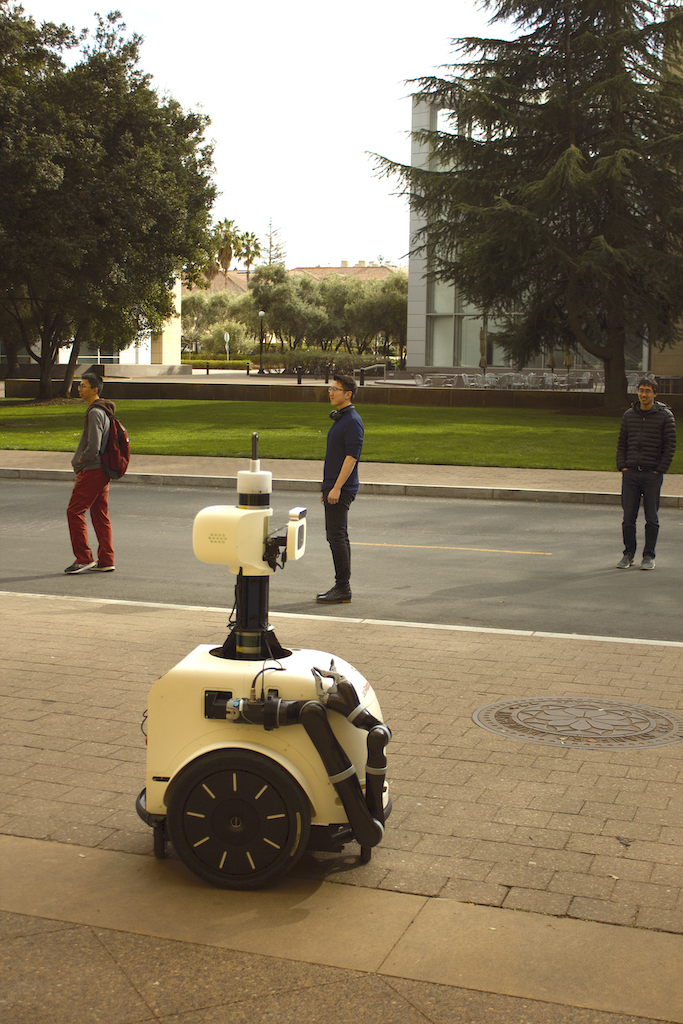}
\end{subfigure}
\centering
\begin{subfigure}{0.15\textwidth}
\includegraphics[width=.99\textwidth]{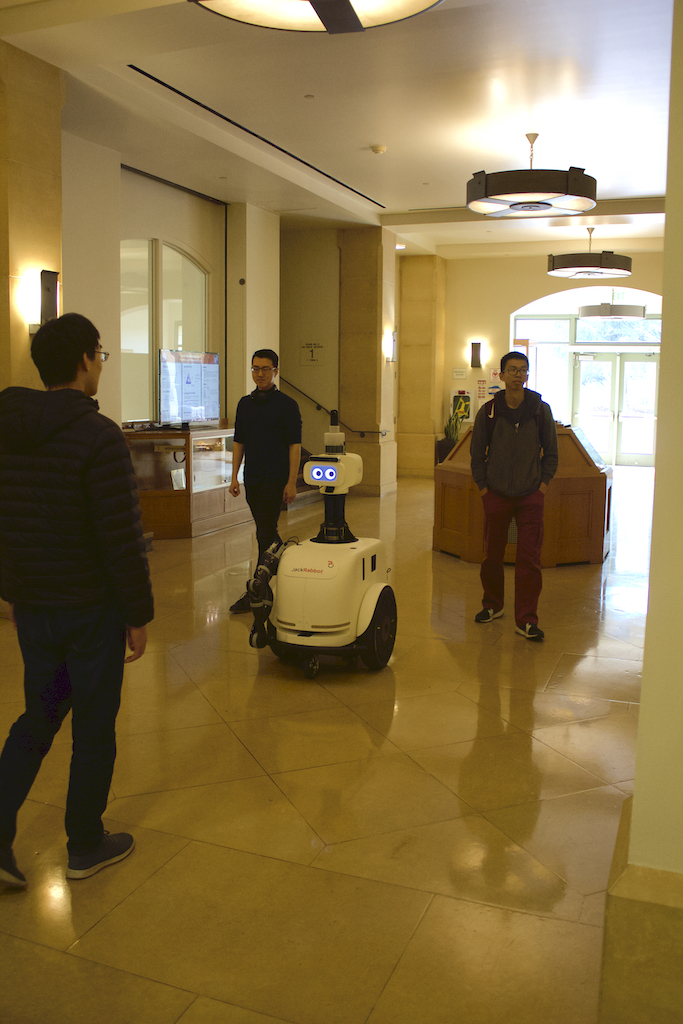}
\end{subfigure}
\begin{subfigure}{0.15\textwidth}
\centering
\includegraphics[width=.99\textwidth]{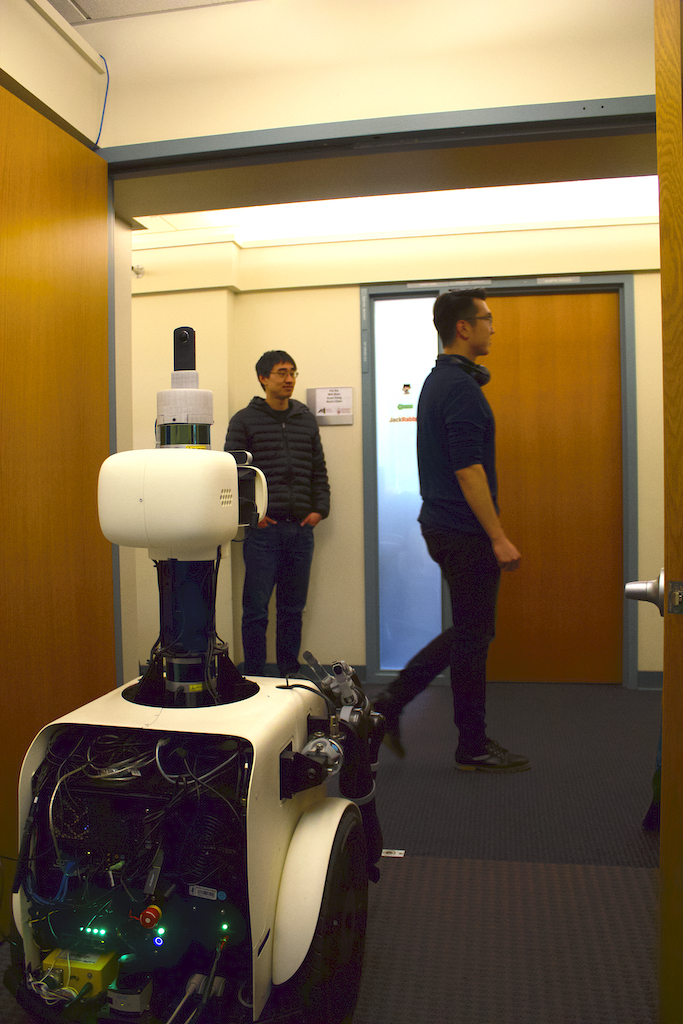}
\end{subfigure}
\caption{We conduct on robot experiments in 3 different scenes, shown above, with a varying number (1 - 7) of people, at different distances (1 - 10m), with different types of human trajectories (moving and stationary), and with \jackrabbot both moving and stationary. We aimed to conduct experiments in diverse, real-world conditions. The above images depict our experimental setup.}
\label{fig:exp_setup}
\end{figure}

We test our solution in three different physical environments, with different lighting conditions (daylight and indoor lighting), with stationary and moving robot, and a different number, distance, and trajectory of moving people. Visualisations of the experimental setup can be seen in Fig.~\ref{fig:exp_setup} We evaluate on a total of 110s of data with 14 unique identities across all scenes. On the on-board computer \methodname runs between 9-11 fps and we measure only 4 ID switches and 1 lost track. These preliminary results, together with the extensive positive results on KITTI and \jrdb, indicate that our tracker provides information to support autonomous navigation in human environments. We make our code publicly available as ROS packages for the community.

%% file: 6_Conclusion.tex
\vspace{-0.5cm}
\section{Conclusion and Future Work}
\label{s:conc}

We presented \methodname, a novel 3D MOT system that fuses the information contained in 2D RGB images and 3D pointclouds in an efficient manner to provide robust tracking performance even in adversarial and highly crowded environments, all while running in real time. 
As part of our project we release the \jrdb dataset, a novel dataset for 2D and 3D MOT evaluation and development containing multimodal data acquired in human environments, including inside university buildings and pedestrian areas on campus, as well as scenes where the robot navigates among humans. 
% The dataset of temporally synchronized and calibrated data includes images from 360\degree stereo cylindrical cameras, LiDAR 3D point clouds, audio signals, IMU values and encoder readings from the robot's base. 
The dataset has been annotated with ground truth 2D bounding boxes and associated 3D cuboids of all persons in the scenes, which will help future research in 2D and 3D MOT. We establish a strong baseline for 3D MOT with \methodname.
\methodname achieves state of the art performance in the well-known KITTI 2D MOT benchmark and shows better performance than existing 3D MOT systems in our provided \jrdb dataset. We also have preliminary on-robot experiments which validate the effectiveness of JRMOT in a real world setting. \methodname serves as a competitive baseline to encourage further research within the paradigm of leveraging multi-modal sensor measurements to better perform 3D MOT. 
% We expect this dataset to support research in perception for autonomous agents in human and social contexts. Our future plans include continue annotating ground truth values for individual and group activities, social grouping, and human posture. 